\documentclass[sigconf,screen,nonacm]{acmart}

\usepackage{algorithm}
\usepackage{algorithmic}
\usepackage{enumitem}
\usepackage{booktabs} 
\usepackage{multirow} 
\usepackage{multicol}
\usepackage{xcolor}  
\usepackage{colortbl} 
\usepackage[most]{tcolorbox}

\usepackage{fancyhdr}
\renewcommand\footnotetextcopyrightpermission[1]{}

\setcopyright{none}
\copyrightyear{2026}
\acmYear{2026}
\acmDOI{XXXXXXX.XXXXXXX}
\acmConference[arXiv preprint]{}{2026}{}

\begin{document}

\title{\textbf{DIRECT: Video Mashup Creation via Hierarchical Multi-Agent Planning and Intent-Guided Editing}}

\settopmatter{authorsperrow=1}

\author{Ke Li\textsuperscript{1},\hspace{1em} Maoliang Li\textsuperscript{2},\hspace{1em} Jialiang Chen\textsuperscript{1},\hspace{1em} Jiayu Chen\textsuperscript{2}, \\ Zihao Zheng\textsuperscript{2},\hspace{1em} Shaoqi Wang\textsuperscript{3},\hspace{1em} Xiang Chen\textsuperscript{2}}
\authornote{Corresponding author. }
\affiliation{%
  \institution{\textsuperscript{1}School of Electronics Engineering and Computer Science, Peking University}
  \institution{\textsuperscript{2}School of Computer Science, Peking University \hspace{0.3em} \textsuperscript{3}Huazhong University of Science and Technology}
  \country{}
}

\renewcommand{\shortauthors}{}








\begin{abstract}
Video mashup creation represents a complex video editing paradigm that recomposes existing footage to craft engaging audio-visual experiences, demanding intricate orchestration across semantic, visual, and auditory dimensions and multiple levels. However, existing automated editing frameworks often overlook the cross-level multimodal orchestration to achieve professional-grade fluidity, resulting in disjointed sequences with abrupt visual transitions and musical misalignment. To address this, we formulate video mashup creation as a Multimodal Coherency Satisfaction Problem (MMCSP) and propose the DIRECT framework. Simulating a professional production pipeline, our hierarchical multi-agent framework decomposes the challenge into three cascade levels: the Screenwriter for source-aware global structural anchoring, the Director for instantiating adaptive editing intent and guidance, and the Editor for intent-guided shot sequence editing with fine-grained optimization. We further introduce Mashup-Bench, a comprehensive benchmark with tailored metrics for visual continuity and auditory alignment. Extensive experiments demonstrate that DIRECT significantly outperforms state-of-the-art baselines in both objective metrics and human subjective evaluation. \\ Project page \& code: \color{magenta}\href{https://github.com/AK-DREAM/DIRECT}{https://github.com/AK-DREAM/DIRECT}
\end{abstract}



\begin{teaserfigure}
  \centering
  \includegraphics[width=\linewidth]{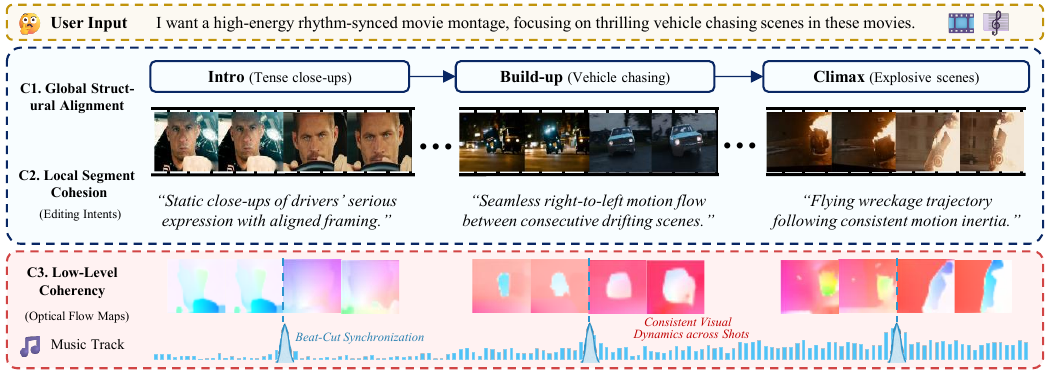}
  \caption{Hierarchical constraints of multimodal coherency in video mashup. (1) \textbf{Global Structural Alignment} of narrative flow, visual elements and musical progression; (2) \textbf{Local Segment Cohesion} through adaptive editing intents that embody multimodal synergy; (3) \textbf{Low-Level Coherency} from fluid visual transitions and precise auditory-visual alignment. }
  \label{fig:teaser}
\end{teaserfigure}

\maketitle


\section{Introduction}

Driven by the growing demand for high-quality multimedia content, video editing represents a vital and complex domain in digital creation. Within this domain, \textbf{Video Mashups}~\cite{wiki:videomashup} (e.g., movie montages and anime music videos) represent a highly demanding creative editing task requiring sophisticated multi-source recomposition. Unlike video summarization tasks~\cite{narasimhan2021clip,hua2025v2xum} that primarily focus on informative content extraction, mashup creation emphasizes aesthetic multimodal orchestration. It aims to select and arrange shots from multiple source videos into a unified sequence where the visual flow is seamlessly continuous and meticulously synchronized with the background music, delivering a coherent and rhythmically engaging viewing experience. To create a high-quality mashup, the editing process involves arranging diverse shots to achieve seamless visual transitions between consecutive scenes (e.g., fluid motion flow and matched framing) and precise musical alignment (e.g., audio-visual intensity matching and beat-cut synchronization)~\cite{murch2001blink}. Nonetheless, the creative task is highly labor-intensive and requires deep editing expertise, motivating the need for automated systems capable of professional-grade mashup creation.

However, existing automated video editing frameworks ~\cite{wang2024lave,sandoval2025editduet,argaw2024towards,ding2025prompt,yang2023shot} struggle to fulfill these creative demands. They primarily follow a semantic-centric paradigm, where video shots are treated as isolated semantic units and selected based on script-guided semantic retrieval. While some methods are music-aware~\cite{videoagent2025, zhu2025weakly}, they mainly focus on simple rhythmic alignment or heuristic music-shot correspondence, neglecting the deeper nuances of dynamic editing pacing and musical progression matching. Furthermore, most approaches ignore fine-grained cues such as motion dynamics and perform no explicit modeling of fluid visual transitions. Crucially, these approaches focus on \textit{which} shots to select, but often overlook \textit{how} the multimodal content should be coherently orchestrated across multiple levels, leading to results that lack the visual continuity and auditory alignment of professional work. 

To systematically overcome these limitations, we draw inspiration from human creative workflows and formulate video mashup creation as a \textbf{Multimodal Coherency Satisfaction Problem} (\textit{MMCSP}). Departing from the conventional semantic-centric paradigm, this formulation defines a unified optimization objective that seeks to jointly satisfy cross-level coherency constraints, spanning both high-level perceptual goals and low-level explicit metrics. As illustrated in Fig.~\ref{fig:teaser}, the \textit{MMCSP} consists of three hierarchical constraints:
(1) \textbf{Global Structural Alignment.} At the global level, a high-quality mashup aligns the visual narrative flow with the musical progression, ensuring a coherent global structure (e.g., matching static close-ups with musical intro v.s. pairing explosive visuals with the climax). 
(2) \textbf{Local Segment Cohesion.} At the local level, cohesive video segments stem from adaptive editing intents that embody multimodal synergy between visual content and editing styles (e.g., rapid cuts and fluid motion for chasing sequence v.s. stable, prolonged shots for emotional scene).
(3) \textbf{Low-Level Coherency.} At the micro level, the shot sequence must satisfy fine-grained coherency constraints measured by explicit metrics, such as visual continuity across transitions and beat-cut synchronization.

The \textit{MMCSP} comprises cross-level heterogeneous constraints that demand a deep synergy between high-level reasoning and low-level optimization, which are inherently difficult to jointly address within a monolithic architecture. This motivates a hierarchical collaborative framework rather than conventional end-to-end models or single-pass LLM prompts.
To this end, we propose \textbf{DIRECT} (\textbf{D}ynamic \textbf{I}ntent for \textbf{R}etrieval \& \textbf{E}diting for \textbf{C}inematic \textbf{T}ransitions), a hierarchical multi-agent framework. Simulating a professional editing pipeline, \textbf{DIRECT} decomposes the complex task into three collaborative modules: 
(1) The \textbf{Screenwriter} acts as the structural architect. To address the challenge of mapping diverse visual elements from the massive, unannotated footage library to the underlying musical structure, it first organizes the available footage into a structured semantic index via semantic clustering and captioning. It then performs music-driven structure anchoring to establish \textit{global structural alignment}, guiding the downstream editing process.
(2) The \textbf{Director} acts as the middleware, synthesizing visual content with dynamic editing styles to formulate adaptive editing intents that embody \textit{local segment cohesion}. These abstract intents are then translated into explicit constraints for precise execution. It also addresses editing failures caused by rigid constraints via a closed-loop validation mechanism with the Editor.
(3) The \textbf{Editor} serves as the executor, performing intent-guided shot retrieval and orchestration to optimize \textit{low-level coherency} metrics. To navigate the combinatorial search space of shot sequences, it employs a constrained path search algorithm, in particular incorporating a dynamic sliding-window trimming mechanism to ensure frame-level precision in both visual continuity and auditory alignment.

In summary, our contributions are as follows:
\begin{itemize}[leftmargin=1.5em, topsep=2pt]
    \item We formulate video mashup creation as a Multimodal Coherency Satisfaction Problem (\textit{MMCSP}). Specifically, we formalize the cross-level coherency constraints and combine them into a unified optimization objective that integrates low-level explicit metrics and high-level perceptual goals.
    \item We introduce \textbf{DIRECT}, a hierarchical multi-agent framework. By integrating hierarchical planning of MLLM agents with fine-grained multimodal perception and optimization, the system effectively solves the \textit{MMCSP} for video mashup.
    \item To address the lack of standardized evaluation for video mashup, we construct \textit{Mashup-Bench}, a new video editing benchmark equipped with metrics specifically designed for visual continuity and auditory synchronization. Extensive experiments demonstrate the effectiveness of our framework in both quantitative metrics and subjective human evaluation.
\end{itemize}
\section{Related Works}

\subsection{Intelligent Video Editing}
Intelligent video editing aims to leverage machine learning techniques to assist in the selection and orchestration of footage.
Early approaches relied predominantly on heuristic rules~\cite{liao2015audeosynth,wang2019write,lee2022popstage} or learning-based models~\cite{liu2023emotion,zhu2025weakly,lu2025skald,yang2023shot,argaw2024towards,chen2025esa}. T2V~\cite{xiong2022transcript} employs shot retrieval and temporal coherence modules to automatically sequence video shots from scripts. Match Cutting~\cite{chen2023match} utilizes metric learning to identify shot pairs with visually coherent transitions. However, these approaches either lack narrative interpretability or rely on manual script planning, limiting their utility for autonomous video creation.
Recently, agentic video editing frameworks~\cite{sandoval2025editduet,videoagent2025,ding2025prompt} such as LAVE~\cite{wang2024lave} leverage LLMs for high-level planning, script generation, and tool orchestration. Despite their success in automated planning and semantic alignment, they largely neglect the intricate multimodal orchestration essential for video mashup.
Our work bridges this gap by integrating hierarchical MLLM planning with fine-grained multimodal perception and optimization for professional video mashup creation.

\subsection{Multi-Agent System}
The emergence of Large Language Models (LLMs) and Multimodal LLMs (MLLMs)~\cite{qwen2025,chen2024internvl} has propelled the development of multi-agent systems~\cite{wu2024autogen,hong2023metagpt}, which autonomously tackle complex problems through collaborative interaction between agents. 
For video creation, recent frameworks employ MLLM-based agents to interpret abstract user intents for script generation~\cite{zhuang2024vlogger,long2024videostudio} or utilize visual capabilities for self-assessment~\cite{yue2025vstylist,yuan2024mora}.
We extend these capabilities to video editing, proposing an MLLM-driven collaborative system that automates complex video mashup tasks.

\subsection{Multimodal Content Retrieval}
A core issue in automated video editing is effectively locating ideal footage from the vast library. Foundational Vision-Language models like CLIP~\cite{radford2021clip} and ImageBind~\cite{girdhar2023imagebind} have revolutionized this by aligning visual and textual representations. Extensions such as CLIP4Clip~\cite{luo2022clip4clip} further adapted these embeddings for the temporal domain, enabling precise video-text matching~\cite{xu2021videoclip,ma2022xclip}.
More recently, frameworks like VideoRAG~\cite{ren2025videorag} leverage graph-based knowledge to enhance long-form video understanding and retrieval. However, these methods predominantly optimize for semantic accruacy, while our method incorporates fine-grained audio-visual features to optimize low-level coherency across shots.

\section{Problem Formulation}

To systematically address video mashup creation as a \textit{MMCSP}, we first characterize the problem space and formalize it by defining a set of sub-objectives comprising both low-level metrics and high-level goals. Finally, we formulate it into a unified optimization objective to guide the automated creation process.

\subsection{Task Definition}
\label{sec:definition}
Given a source video library $\mathcal{S}$, a background music track $\mathcal{M}$, and user instruction $\mathcal{I}$, our goal is to recompose a sequence of visual shots aligned with $\mathcal{M}$ to generate a final video mashup $\mathcal{V}=\{v_i\}_{i=1}^m$ that not only satisfies the user's instruction but also exhibits professional-grade multimodal coherency.

\subsection{Low-Level Metrics}
\label{sec:quant_metrics}
To explicitly measure the low-level coherency requirements, we define six quantifiable metrics across three multimodal dimensions.

\vspace{2pt} 
\noindent \textbf{Semantic - Prompt Relevance ($m_1$).}
This metric measures the semantic relevance between a selected shot $v_i$ and the user prompt $\mathcal{I}$. Using a vision-language representation model (e.g., CLIP~\cite{radford2021clip}), we calculate the cosine similarity between the semantic embedding of the shot and the user prompt (refined by LLM): $m_1(v_i, \mathcal{I}) = \cos(E(v_i), E(\mathcal{I}))$. Following CLIP4Clip~\cite{luo2022clip4clip}, we derive the shot embedding $E(v_i)$ by averaging the embeddings of its constitutive keyframes; this aggregation strategy is adopted by default throughout this work unless otherwise specified.

\vspace{2pt} 
\noindent \textbf{Semantic - Segment Consistency ($m_2$).}
Frequent jumps in visual semantics can cause audience confusion and an incoherent narrative experience; therefore, we measure segment semantic consistency by computing the cosine similarity between visual embeddings of consecutive shots: $m_2(v_{i-1}, v_i) = \cos(E(v_{i-1}), E(v_i))$.

\vspace{2pt} 
\noindent \textbf{Visual - Motion Continuity ($m_3$).}
Professional editing achieves fluid visual transitions between consecutive shots by matching the visual motion direction and velocity. To measure motion continuity, we calculate the optical flow field similarity between consecutive shots: $m_3(v_{i-1}, v_i) = \text{sim}(F_{end}(v_{i-1}), F_{start}(v_i))$, where $F(\cdot)$ represents the optical flow field at the start/end of the shot.

\vspace{2pt} 
\noindent \textbf{Visual - Framing Consistency ($m_4$).} Another professional technique is the use of match-cutting~\cite{chen2023match} to overlap subjects' positions between consecutive shots, providing a sense of framing consistency and preventing erratic focal point jumps. We quantify this by computing the similarity between the foreground saliency maps at the shot boundary: $m_4(v_{i-1}, v_i) = \text{sim}(S_{end}(v_{i-1}), S_{start}(v_i))$, where $S(\cdot)$ denotes the saliency map at the start/end of the shot.

\vspace{2pt} 
\noindent \textbf{Auditory - Beat-Cut Synchronization ($m_5$).} Precise temporal synchronization between visual cuts and musical beats is essential for an immersive rhythmic mashup. We calculate the sync score based on the temporal proximity between the shot boundary and its closest musical beat: $m_5(v_i, \mathcal{M}) = \exp(-||T_{cut}-T_{beat}||)$.

\vspace{2pt} 
\noindent \textbf{Auditory - Energy Correspondence ($m_6$).} A compelling mashup requires a precise correspondence between the musical intensity and the visual dynamics. We quantify this by calculating the Spearman correlation~\cite{spearman} between the average optical flow magnitude of the visual shot and the root-mean-square (RMS) energy of the music track: $m_6(\mathcal{V}, \mathcal{M}) = \text{corr}(||F(v_i)||_{i=1}^m,\text{RMS}(\mathcal{M}))$.

Our detailed implementation specifications of these metrics are available in the Supplementary Material.

\begin{figure*}[ht!] 
  \centering
  \includegraphics[width=\textwidth]{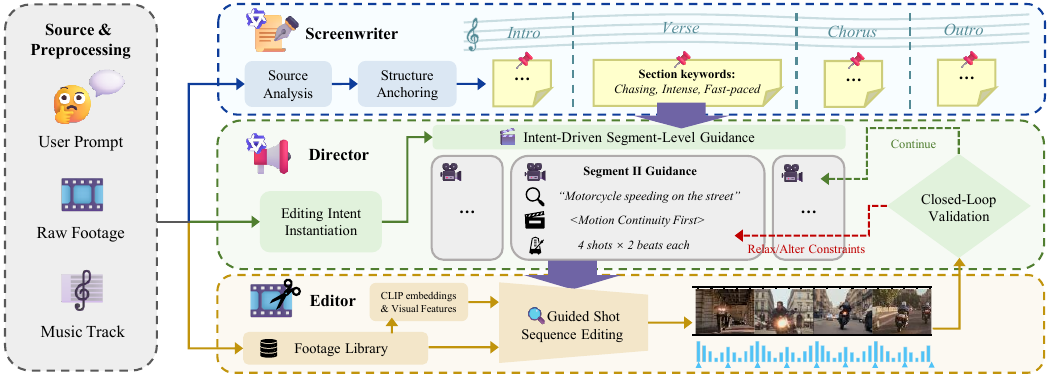}
  \caption{Overview of DIRECT. We decompose video mashup creation into three collaborative modules: the Screenwriter anchors the global structure to align the multimodal content; the Director instantiates segment-level guidance (query, heuristic, pacing); and the Editor executes shot retrieval and orchestration following the editing guidance with closed-loop validation.}
  \label{fig:framework}
\end{figure*}

\subsection{High-Level Goals}
\label{sec:high_level_obj}
While the explicit metrics represent low-level coherency, relying solely on them often results in a "hollow" video—a collection of well-matched but disjointed shots lacking a holistic structure and local segment cohesion. To address this, we introduce two high-level goals that reflect human perceptual quality of professional-grade video mashup, in a hierarchical manner of \textit{global structure alignment} and \textit{local segment cohesion}.

\vspace{2pt} 
\noindent \textbf{Global Structural Alignment ($G_{global}$).}
This objective defines the multimodal structural alignment of the mashup, where the narrative flow, visual elements, and musical progression are orchestrated along a unified timeline to form a \textbf{coherent global structure}.
As shown in Fig.~\ref{fig:teaser}, a vehicle chasing mashup satisfies global structural alignment by matching static driver close-ups with the musical intro to build tension, while pairing high-energy vehicle explosions with the climax. 
It ensures that the mashup possesses a well-organized structure, rather than being a disordered collection of shots.

\vspace{2pt} 
\noindent \textbf{Local Segment Cohesion ($G_{local}$).}
This objective measures the segment-level aesthetics of the mashup, which extends beyond optimizing isolated metrics. It requires the joint consideration of local multimodal aspects, where shot semantics, visual editing heuristics, and rhythmic pacing are synthesized to reflect a \textbf{high-level editing intent}. As shown in the middle column of Fig.~\ref{fig:teaser}, a fluid drifting scene transition prioritizes seamless motion flow and precise beat-cut alignment. In contrast, an emotional scene demands semantically precise and stable, prolonged takes.
It ensures that local segments exhibit intent-guided multimodal synergy rather than merely maximizing explicit metrics.

\subsection{Unified Optimization Objective}
\label{sec:unified_goal}

Combining the low-level explicit metrics and high-level goals, we formulate the \textit{MMCSP} for video mashup creation as a joint optimization problem. The optimal video result $\mathcal{V}^*$ is defined as:
\begin{equation}
\label{eq:unified_goal}
    \mathcal{V}^* = \operatorname*{argmax}_{\mathcal{V}} \left[ G_{global}(\mathcal{V}) + G_{local}(\mathcal{V}) + \sum_{i=1}^t w_i \cdot M_i(\mathcal{V}) \right]
\end{equation}
where $M_i(\mathcal{V})$ denotes the average score of the $i$-th explicit metric $m_i$ across all shots and $w_i$ are the balancing weights for them.

Directly solving Eq.~\ref{eq:unified_goal} presents a fundamental challenge. While the explicit metrics ($M_i$) are quantifiable, the high-level goals ($G_{global}$ and $G_{local}$) rely on structural design and editing intuition. This highlights that \textit{MMCSP} is not merely a metric maximization problem, but a complex task that requires a deep synergy between high-level agentic planning and low-level algorithmic optimization.
\begin{figure*}[ht!]
  \centering
  \includegraphics[width=\textwidth]{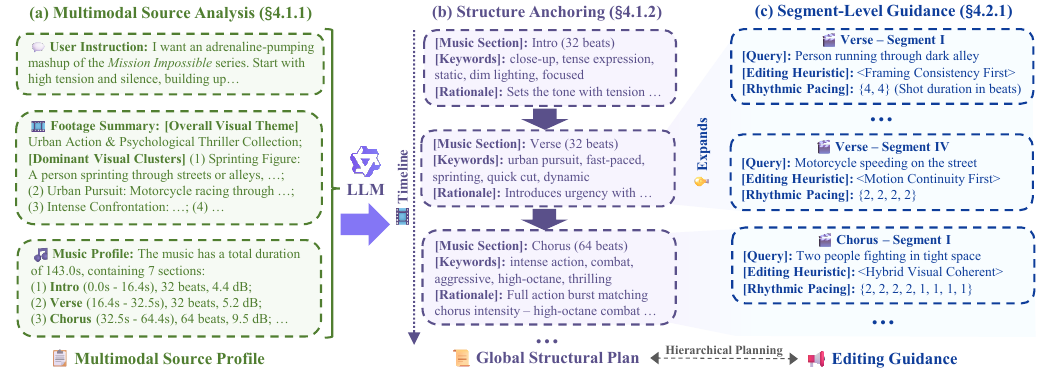}
  \caption{Visualization of the hierarchical planning workflow in DIRECT. The Screenwriter leverages multimodal source analysis to generate a section-wise global structural plan, and the Director expands it into segment-level editing guidance. }
  \label{fig:planning}
\end{figure*}

\section{Methodology}

To solve the joint optimization problem above, we propose the \textbf{DIRECT} framework. 
As illustrated in Fig.~\ref{fig:framework}, the framework operates through three collaborative modules: the \textbf{Screenwriter} for anchoring global structural alignment ($G_{global}$), the \textbf{Director} for instantiating local segment cohesion ($G_{local}$), and the \textbf{Editor} for optimizing low-level coherency ($M_i$).

Given the raw footage, we first construct a fine-grained footage library $\mathcal{L}$ through an automated preprocessing pipeline. Specifically, we segment raw footage into discrete shots and employ established models to extract frame-level features, including cross-modal semantic embeddings~\cite{radford2021clip}, saliency maps~\cite{qin2020u2}, and optical flow~\cite{teed2020raft}. This pre-computation facilitates subsequent footage summarization and enables inference-free metric evaluation of candidate sequences during the sequence editing phase.
\subsection{Screenwriter: Global Alignment Anchoring}
\label{sec:screenwriter}
The \textbf{Screenwriter} is designed to tackle the global structural alignment goal ($G_{global}$), which is inherently intractable for direct algorithmic optimization and demands high-level reasoning.
As shown in Fig.~\ref{fig:planning}(a,b), by leveraging multimodal source analysis and music-driven structure anchoring, it constructs a coherent global structural plan that anchors the narrative arc and available visual footage to align the temporal musical structure. 

\vspace{2pt} 
\noindent \textbf{Multimodal Source Analysis.}
To ensure that the generated global structure is grounded in source assets, the Screenwriter requires a comprehensive understanding of the multimodal source content. For visual footage, directly feeding the Screenwriter with thousands of raw shots is excessively costly and triggers context overflow. Instead, we propose an effective \textbf{footage summarization} strategy based on semantic clustering.
We first group all shots in the footage library $\mathcal{L}$ into a set of clusters $\mathcal{C}$ based on their preprocessed semantic embeddings. 
For each cluster $c\in \mathcal{C}$, an MLLM inspects its representative shots $\mu(c)$ (i.e., cluster centroids) to extract a caption of the shared visual subject. The captions of each cluster are then synthesized by LLM into a structured \textit{Footage Summary} $\mathcal{F}_{sum}$ consisting of the library's overall visual theme and dominant visual clusters (e.g., "urban nightscapes" or "high-speed vehicle chase"). The process can be defined as follows:
\begin{equation}
    \mathcal{F}_{sum} = \text{Synth} (\{\text{MLLM}\left(\mu(c), \mathcal{I}_{cap} \right)\}_{c \in \mathcal{C}}, \mathcal{I}_{syn}),
\end{equation}
where $\mathcal{I}_{cap},\mathcal{I}_{syn}$ are the instructions for visual captioning and summary synthesis, respectively.
For the music track $\mathcal{M}$, we utilize an off-the-shelf music analysis model~\cite{kim2023all} to extract its temporal structure, yielding a \textit{Music Profile} $\mathcal{M}_{stru}$ that maps $N$ musical sections (e.g., intro, chorus) and their intensity to the respective timestamps. It delineates the musical progression, providing a temporal backbone for global structural alignment.

\vspace{2pt} 
\noindent \textbf{Music-Driven Structure Anchoring.}
The core challenge of constructing a coherent global structure is the temporal alignment of the multimodal content. To address this, we propose a music-driven approach that generates a section-wise plan according to the musical structure.
Specifically, the Screenwriter synthesizes the footage summary $\mathcal{F}_{sum}$ and music profile $\mathcal{M}_{stru}$ to generate a global structural plan $\mathcal{K}$ that aligns multiple modalities:
\begin{equation}
    \mathcal{K} = \{K_i\}_{i=1}^N = \text{LLM}(\mathcal{F}_{sum}, \mathcal{M}_{stru}, \mathcal{I}_{user},\mathcal{I}_{plan}),
\end{equation}
where $I_{user}$ represents user instruction and $I_{plan}$ denotes the system prompt. Each $K_i$ consists of a set of stylistic or descriptive \textbf{section keywords} (e.g., fast-paced, intense combat) that anchor the narrative flow and intended visual elements to the i-th musical section. The section-wise keywords bridge the semantic, visual, and auditory modalities for global structural alignment, providing an explicit plan for the downstream intra-section editing.

\subsection{Director: Local Cohesion Instantiation}
\label{sec:director}

With the structural plan established, the system needs to instantiate the abstract plan into a concrete shot sequence. However, simply maximizing explicit metrics in Sec.~\ref{sec:quant_metrics} often fails to ensure high-level segment cohesion ($G_{local}$) due to the lack of multimodal-aware editing intent. 
To bridge this gap, the \textbf{Director} instantiates concrete editing intents from the abstract section keywords, interacting with the Editor through a closed-loop workflow comprising three phases: \textit{Guidance}, \textit{Editing}, and \textit{Validation}.

\vspace{2pt} 
\noindent \textbf{Intent-Driven Segment-Level Guidance.}
As shown in Fig.~\ref{fig:planning}(c), to enable high-level intent guided editing, the Director partitions each musical section into shorter \textbf{segments} and generates a detailed editing guidance for each segment, including three constraints:
(1) \textit{Semantic Query} ($q_{sem}$): It expands abstract section keywords into a concrete visual subject for this segment (e.g., motorcycle speeding on the street), filtering the footage library to construct a semantically relevant shot pool.
(2) \textit{Editing Heuristic} ($\mathbf{w}_{heu}$): The Director selects an optimal heuristic from predefined templates (e.g., \textit{Motion-Continuity-First} vs. \textit{Semantic-Precision-First}). This dynamically assigns values to the low-level metrics' balancing weights in Eq.~\ref{eq:unified_goal}, ensuring that the Editor prioritizes metrics that align with the specific editing intent.
(3) \textit{Rhythmic Pacing} ($\tau_{pace}$): It defines the temporal structure for the segment (e.g., four shots with a 2-beat duration each), ensuring precise beat-cut alignment by matching shot boundaries to musical beats, while modulating the editing pace to reflect the editing intent.
Formally, the Director expands the keywords $K_i$ of each musical section into a sequence of segment-level editing guidance $\{\mathcal{G}_{i,1}, \dots, \mathcal{G}_{i,M}\}$, defined as:
\begin{equation}
    \mathcal{G}_{i,j} = \text{Director}(K_i, \mathcal{F}_{sum}, \dots) = \langle q_{sem}, \mathbf{w}_{heu}, \tau_{pace} \rangle_{i,j},
\end{equation}

\vspace{2pt} 
\noindent \textbf{Guided Shot Sequence Editing.}
Given the editing guidance for a segment, the Editor executes a constrained search within the footage library. It aims to identify candidate shot sequences that satisfy the semantic query $q_{sem}$ and the rhythmic pacing $\tau_{pace}$, while optimizing the objective function with respect to the heuristic weights $\mathbf{w}_{heu}$. The detailed procedure is elaborated in Sec.~\ref{sec:editor}.

\vspace{2pt} 
\noindent \textbf{Closed-Loop Validation.} A non-trivial challenge in automatic editing is that the Editor may fail to produce shot sequences that align with the Director's guidance. Although grounded in the footage summary $\mathcal{F}_{sum}$, overly rigid semantic queries ($q_{sem}$) can lead to limited relevant shots, resulting in suboptimal or mismatched sequences. To address this, the Director employs an MLLM-based validator to assess the candidate shot sequences returned by the Editor. If all candidates fail to meet the guidance criteria, the Director dynamically relaxes or alters the semantic query based on validation feedback (e.g., generalizing "man sprinting through dark alley" to "man running fast"). This adaptively expands or shifts the relevant shot pool, allowing the system to escape local optima caused by insufficient shots. This process repeats until a satisfactory sequence is selected or the maximum retry limit is reached.
\subsection{Editor: Intent-Guided Sequence Editing}
\label{sec:editor}

Although LLMs excel at high-level planning, they lack the fine-grained multimodal perception required for low-level coherency. Consequently, the \textbf{Editor} executes the Director's editing guidance by leveraging pre-extracted features to orchestrate shot sequences that maximize the explicit metrics defined in Eq.~\ref{eq:unified_goal} (i.e., $M_i$). 

We formulate the task as a constrained path search problem within a footage graph $\mathcal{G} = (\mathcal{V}_{pool}, \mathcal{E})$, where each node $v\in \mathcal{V}_{pool}$ represents a raw shot in the relevant shot pool filtered by $q_{sem}$, and an edge $e\in \mathcal{E}$ denotes a transition between sequential shots. Any valid path $\mathcal{P}=\{\hat{v}_1,\dots,\hat{v}_K\} \in \Omega$ constitutes a shot sequence, where each $\hat{v}_i$ is a sub-clip trimmed from the raw shot $v_i$ to fulfill the $\tau_{pace}$ constraint and $\Omega$ denotes the set of all valid sequences. The goal is to find the optimal shot sequence $\mathcal{P}^*$ that maximizes the explicit metrics with balancing weights $\mathbf{w}_{heu}$:

\begin{figure}[h]
\centering
  \includegraphics[width=\linewidth]{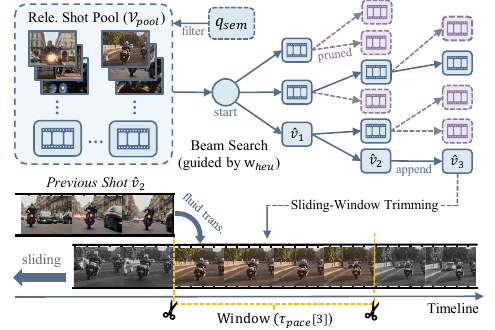}
  \caption{Intent-Guided Shot Sequence Editing. The Editor uses a tailored beam search algorithm with dynamic sliding-window trimming to find optimal shot sequences.}
  \label{fig:editing}
\end{figure}

\begin{equation} 
    \label{eq:beam_search_goal}
    \mathcal{P}^* = \operatorname*{argmax}_{\mathcal{P}\in \Omega} [\text{Score}(\mathbf{w}_{heu},\mathcal{P})] =\operatorname*{argmax}_{\mathcal{P}\in \Omega}  [\sum_i \mathbf{w}^{heu}_i\cdot M_i(\mathcal{P})]
\end{equation}

As shown in Fig.~\ref{fig:editing}, we propose a tailored beam search algorithm to efficiently identify optimal shot sequences within the vast search space. Initially, we filter the footage library $\mathcal{L}$ to construct a relevant shot pool $\mathcal{V}_{pool}$ based on the cosine similarity of the cross-modal embeddings between $q_{sem}$ and all shots. During the $k$-th iteration of the beam search, we expand all current partial sequences by appending candidate shots from $\mathcal{V}_{pool}$ using a sliding-window trimming mechanism. Specifically, the algorithm enumerates all temporal windows within the appended raw shot that satisfy the duration constraint of current position ($\tau_{pace}[k]$) to maximize the composite score in Eq.~\ref{eq:beam_search_goal}. This dynamically extracts the optimal sub-clip for low-level coherency metrics; in particular, it calibrates the cut-point to maximize inter-shot visual continuity ($m_3,m_4$), ensuring a fluid visual transition with the previous shot. Subsequently, the algorithm prunes the search tree and retains only the top-$B$ sequences with the highest cumulative scores for the next iteration. Finally, the $B$ best complete sequences are returned to the Director for validation. Notably, this process remains computationally efficient as all frame-level features are pre-computed in the footage library $\mathcal{L}$, enabling rapid score evaluation without real-time inference.
\section{Experiments}

\begin{table*}[ht!]
    \centering
    \caption{Main Results on Baseline Comparison and Ablation Study. Results are averaged over all test cases for quantitative metrics $m_1$--$m_6$ (scaled by 100) and a 10-cases subset for human evaluation. Best in bold, second best \underline{underlined}.}
    \label{tab:main_results}
    \begin{tabular}{l cc cc cc ccc}
        \toprule
        & \multicolumn{2}{c}{\textbf{Semantic Rele.}} & \multicolumn{2}{c}{\textbf{Visual Cont.}} & \multicolumn{2}{c}{\textbf{Auditory Align.}} & \multicolumn{3}{c}{\textbf{Human Eval.}} \\
        \cmidrule(lr){2-3} \cmidrule(lr){4-5} \cmidrule(lr){6-7} \cmidrule(lr){8-10}
        \textbf{Method} & \textbf{Prom.}($m_1$)$\uparrow$ & \textbf{Seg.}($m_2$)$\uparrow$ & \textbf{Mot.}($m_3$)$\uparrow$ & \textbf{Fram.}($m_4$)$\uparrow$ & \textbf{Sync.}($m_5$)$\uparrow$ & \textbf{Ener.}($m_6$)$\uparrow$ & \textbf{Glob.}$\uparrow$ & \textbf{Loc.}$\uparrow$ & \textbf{Qual.}$\uparrow$\\
        \midrule
        T2V*~\cite{xiong2022transcript} & \underline{25.89} & 81.80 & 61.72 & 76.21 & 91.46 & 52.67 & 3.6 & 2.6 & 2.7 \\
        MMSC~\cite{zhu2025weakly} & 25.63 & \textbf{90.24} & 61.64 & 76.43 & 88.25 & 48.23 & 3.3 & 3.5 & 3.1 \\
        VideoAgent~\cite{videoagent2025} & 25.09 & 82.59 & 62.24 & 78.85 & 88.43 & 51.88 & 5.1 & 3.8 & 4.6 \\
        \midrule
        \textit{w/o Screenwriter} & \textbf{25.96} & 84.80 & \underline{76.93} & \underline{83.35} & \underline{98.62} & \textbf{82.41} & 4.4 & \underline{4.8} & \underline{5.0} \\ 
        \textit{w/o Director}     & 25.30 & 84.73 & 75.38 & 83.20 & 98.34 & 82.29 & \underline{5.3} & 3.5 & 4.4 \\
        \textit{w/o dynamic trimming}     & 25.25 & 84.57 & 71.03 & 81.73 & 98.67 & 78.67 & - & - & - \\
        \midrule
        \rowcolor{gray!10}
        \textbf{DIRECT (Full)}      & 25.49 & \underline{84.97} & \textbf{77.26} & \textbf{83.37} & \textbf{98.69} & \underline{82.40} & \textbf{5.8} & \textbf{5.3} & \textbf{5.9} \\
        \bottomrule
    \end{tabular}
\end{table*}

\subsection{Experimental Setup}

\noindent \textbf{Dataset.}
To address the lack of benchmarks specifically designed to evaluate multimodal coherency in video mashups, we introduce \textit{Mashup-Bench}.
The dataset features a diverse footage library spanning five genres sourced from 15 iconic movie series, totaling 38 videos, 4,000 minutes of footage, and 64,000 atomic shots. Complementing the video data, we also select 10 music tracks ranging from 2 to 4 minutes with varying styles and tempos. These assets are paired with human-curated instructions to construct test cases defined in Sec.~\ref{sec:definition}. To assess the system's capability in handling heterogeneous visual distributions, we also introduce cross-series test cases that integrate source footage from multiple series. Aligned with the experimental scale in recent long-form video editing works~\cite{zhu2025weakly,sandoval2025editduet}, we curated 40 test cases in total, comprising over 3,000 orchestrated shots for a comprehensive evaluation.


\begin{figure*}[t] 
  \centering
  \includegraphics[width=\textwidth]{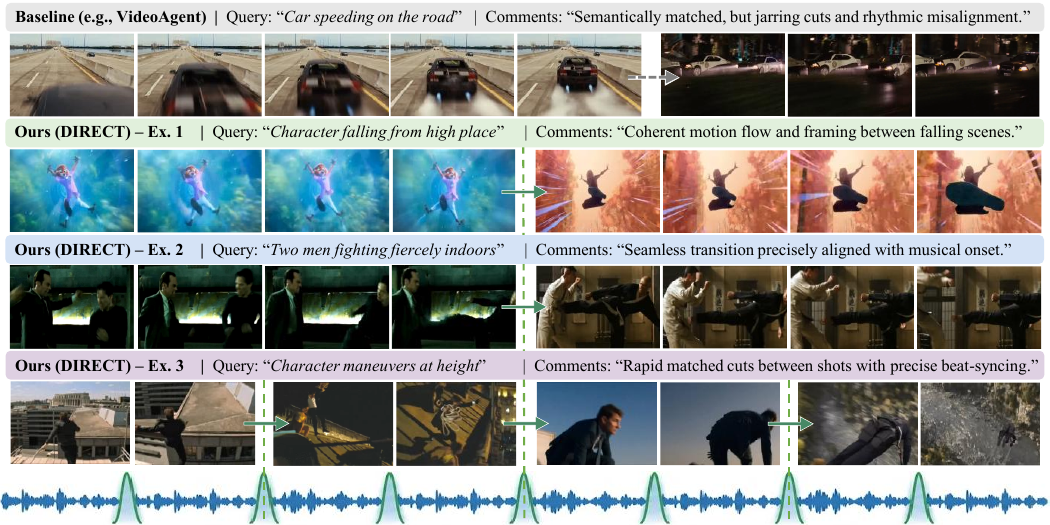}
  \caption{Qualitative Comparison of Low-Level Coherency. While baseline (top row) only ensures semantic relevance, our method achieves superior visual continuity (matched subject position and motion flow across transitions) and auditory alignment (visual cut points synchronized with musical beats indicated by green crests). }
  \label{fig:qual_examples}
\end{figure*}

\vspace{2pt}
\noindent \textbf{Evaluation Metrics.}
To quantitatively assess the quality of the video results, we evaluate the six metrics as detailed in Sec.~\ref{sec:quant_metrics}: \textit{Prompt Relevance} ($m_1$), \textit{Segment Consistency} ($m_2$), \textit{Motion Continuity} ($m_3$), \textit{Framing Consistency} ($m_4$), \textit{Beat-Cut Synchronization} ($m_5$), and \textit{Energy Correspondence} ($m_6$). These metrics provide a multi-dimensional assessment of multimodal coherency across semantic relevance, visual continuity, and auditory alignment. To further assess high-level perceptual goals in Sec.~\ref{sec:high_level_obj} that are difficult to quantify, we also conducted a human study on video results generated by different methods, as elaborated in Sec.~\ref{sec:subj_eval}.

\vspace{2pt} 
\noindent \textbf{Implementation Details.}
In the preprocessing phase, we utilize established models to extract frame-level video features: CLIP (ViT-B/32)~\cite{radford2021clip} for semantic embeddings, RAFT-Large~\cite{teed2020raft} for optical flow, and U2-Net~\cite{qin2020u2} for saliency map. To balance precision with efficiency, we adopt a temporal stride of $S=4$ frames and apply spatial average pooling to compress feature dimensions. Furthermore, we employ PySceneDetect~\cite{pyscenedetect} to segment raw footage into atomic shots, and utilize All-In-One~\cite{kim2023all}, a unified music analyze model to extract musical structures and beat onsets. For all MLLM agents, we deploy open-source Qwen3-VL-8B-Instruct~\cite{qwen2025} under default settings of the vLLM framework~\cite{kwon2023efficient} as the unified backbone. We observe that it exhibited comparable ability to larger models under our hierarchical task decomposition, and choose it in favor of reproducibility and efficient local deployment. For the Editor's hyper-parameters, we set the beam width $B=3$ and the sliding-window stride $S=4$ frames. All experiments are conducted on a local server equipped with four NVIDIA L20 (48GB) GPUs, where the one-time preprocessing for a 2-hour movie requires approximately 25 minutes. During inference, our framework generates a complete video mashup in an average of 10 minutes.

\subsection{Objective Evaluation}
\label{sec:obj_eval}
We compare \textbf{DIRECT} with three representative categories of baselines:
(1) \textbf{Transcript2Video} (T2V)~\cite{xiong2022transcript} is a standard retrieval-based method that aligns shot sequences with textual scripts. Due to the original model being unavailable, we adapt CLIP~\cite{radford2021clip} as its VLM backbone and utilize LLM to generate scripts based on input prompts. 
(2) \textbf{MMSC}~\cite{zhu2025weakly} is a state-of-the-art end-to-end framework for movie trailer/montage generation, using a weakly-supervised learning framework to specifically optimize semantic consistency.
(3) \textbf{VideoAgent}~\cite{videoagent2025} is a state-of-the-art open-source agentic framework for video editing (including movie mashups), involving LLM-based planning for narrative generation and a graph orchestration mechanism managing specialized agents and editing tools. 

As shown in Table~\ref{tab:main_results}, while all methods achieve comparable performance in Semantic Relevance (MMSC highest in $m_2$), \textbf{DIRECT} significantly outperforms all baselines in both Visual Continuity and Auditory Alignment. 
This confirms that while semantic-centric retrieval (used by baselines) only ensures semantic accuracy, our method effectively leverages fine-grained features aware shot retrieval and orchestration by the \textbf{Editor} to achieve superior performance in low-level coherency metrics. 

To provide an intuitive understanding of these quantitative gains, Fig.~\ref{fig:qual_examples} visualizes the low-level coherency. As shown, while the baseline retrieves semantically relevant shots (e.g., car speeding scenes in the top row), they often exhibit jarring visual cuts and rhythmic misalignment. In contrast, \textbf{DIRECT} orchestrates cohesive segments with fluid visual transitions and precise musical synchronization, reflecting the substantial improvements observed in visual continuity and auditory alignment metrics.

\subsection{Subjective Evaluation}
\label{sec:subj_eval}

Complemental to the objective evaluation, we conducted a blind user study to assess the perceptual quality of the generated videos. 
Due to the labor-intensive nature of human evaluation, we randomly sampled a subset of 10 diverse test cases from the benchmark, covering various themes and musical styles. For each test case, we generated videos using \textbf{DIRECT} and the three baselines.
We invited 60 participants (primarily university students from diverse majors) to watch the anonymized videos in random order. 

Following standard human study protocols in previous works~\cite{zhu2025paper2video,zhu2025weakly}, participants were asked to rate each video on a 1-7 Likert scale based on the high-level goals defined in Sec.~\ref{sec:high_level_obj}:
(1) \textbf{Global Structural Alignment} ($G_{global}$): Does the video exhibit a cohesive global structure that aligns the narrative flow and visual elements with the musical progression?
(2) \textbf{Local Segment Cohesion} ($G_{local}$): Are the visual transitions fluid, and does the synergy between shot semantics and editing styles reflect professional-grade aesthetics?
(3) \textbf{Overall Quality:} Does the mashup provide an overall compelling and professional viewing experience? The detailed human evaluation protocol is available in the Supp.\ Mat.

The comparative results are presented in the rightmost columns of Table~\ref{tab:main_results}.
\textbf{DIRECT} outperforms all baselines by a significant margin across all three dimensions. 
Most notably, our method achieves an average score of \textbf{5.3} in Local Segment Cohesion, surpassing the best baseline (VideoAgent) by 40\%.
Crucially, the superiority in Global Structural Alignment and Local Segment Cohesion validates the effectiveness of our hierarchical multi-agent planning workflow, highlighting the pivotal role of \textbf{Screenwriter} and \textbf{Director} agents in achieving the high-level goals.

\begin{figure}[t!]
\centering
  \includegraphics[width=\linewidth]{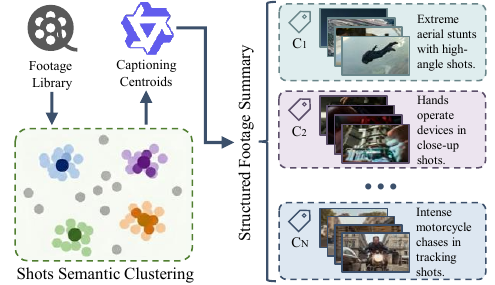}
  \caption{Case study of Footage Summarization. It deconstructs the expansive footage library by clustering and captioning semantically related shots into distinct groups.}
  \label{fig:footage_summary}
\end{figure}

\subsection{Ablation Study}
\label{sec:ablation}
To further verify the necessity of each module, we compare the full \textbf{DIRECT} framework with three variants: 
(1) \textbf{w/o Screenwriter:} The system bypasses the global structure anchoring phase; instead, the Director generates segment-level guidance directly from the raw user prompt.
(2) \textbf{w/o Director:} The system directly retrieves shots based on the structural plan generated by the Screenwriter, using a fixed editing guidance for each segment. 
(3) \textbf{w/o dynamic trimming:} The Editor omits dynamic sliding-window trimming, satisfying pacing constraints by simply truncating shots from their start instead of searching for an optimal temporal window.
The results are reported in the bottom section of Table~\ref{tab:main_results}. 

\vspace{2pt}
\noindent \textbf{Impact of Hierarchical Agents.}
As shown in Table~\ref{tab:main_results}, removing the Screenwriter or the Director yields a negligible difference in low-level metrics ($m_1$--$m_6$). This is within expectation, as the Editor module remains responsible for maximizing the metrics. However, a significant performance gap emerges in subjective scores: excluding the \textbf{Screenwriter} leads to a sharp decline in \textit{Global Structural Alignment} (Glob.), as the generated videos fail to exhibit a coherent narrative flow that aligns with the musical structure without a global plan. Similarly, the absence of the \textbf{Director} results in a notable drop in \textit{Local Segment Cohesion} (Loc.). The system lacks the adaptive editing heuristic and rhythmic pacing for different visual content and musical vibes as it reverts to a fixed segment guidance profile, confirming the Director's vital role in translating high-level editing intent into precise editing guidance.

\vspace{2pt}
\noindent \textbf{Impact of Dynamic Sliding-Window Trimming.} Omitting the dynamic trimming mechanism yields a significant performance drop in low-level coherency metrics, specifically in \textit{Motion Continuity} ($m_3$) and \textit{Framing Consistency} ($m_4$). Without this mechanism, the Editor cannot precisely calibrate cut-points with frame-level precision to align motion flow and subject positions between consecutive shots. This highlights the mechanism's critical role in achieving seamless visual transitions through fine-grained temporal refinement while maintaining precise rhythmic alignment.

\begin{figure}[t!]
\centering
  \includegraphics[width=\linewidth]{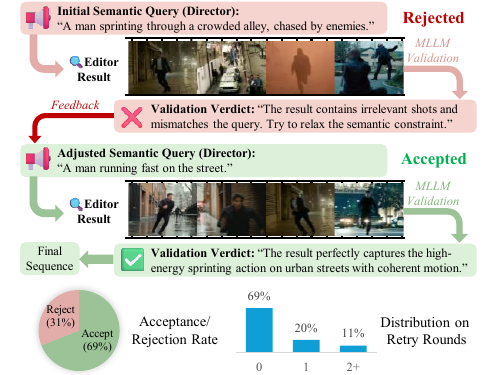}
  \caption{Effectiveness of Closed-Loop Validation. The validator detects editing failures (31\% rejection rate) and prompts query adjustment to ensure sequence coherence.}
  \label{fig:closed_loop}
\end{figure}

\vspace{2pt}
\noindent \textbf{Effectiveness of Footage Summarization.}
Fig.~\ref{fig:footage_summary} demonstrates how our footage summarization mechanism organizes raw shots into a structured footage summary. In the example, raw shots from an action movie are grouped into visual clusters such as "aerial stunt" and "motorcycle chase". Instead of overwhelming MLLM agents with excessive raw shots, this approach provides a structured semantic index, ensuring both processing efficiency and that the subsequent planning is grounded in the available visual footage.

\vspace{2pt}
\noindent \textbf{Effectiveness of Closed-Loop Validation.}
Fig.~\ref{fig:closed_loop} demonstrates our system's ability to resolve editing failures via closed-loop validation. In the example, the initial rigid query yields suboptimal result due to insufficient relevant shots in the footage library. The Director's MLLM-based validator identifies this failure and dynamically refines the query to expand the relevant shot pool. The 31\% overall rejection rate confirms this mechanism's vital role in identifying suboptimal results and adjusting query to find high-quality alternatives while still adhering to the global structural plan. 
\section{Conclusion}

In conclusion, we propose \textbf{DIRECT}, a hierarchical multi-agent framework that reformulates video mashup creation as a Multimodal Coherency Satisfaction Problem \textit{(MMCSP)}. By decomposing the complex editing task into Screenwriter, Director, and Editor modules, DIRECT bridges high-level structural plan and editing intent with low-level coherency aware editing. To evaluate this, we construct \textit{Mashup-Bench}, a comprehensive benchmark specifically designed to assess multimodal coherency. Extensive experiments demonstrate DIRECT's superior ability to generate professional-grade mashups compared to existing baselines. In future work, we plan to extend the \textit{MMCSP} formulation and our framework to broader video editing tasks, focusing on the intent-guided retrieval and orchestration of diverse multimodal assets.

\balance

\bibliographystyle{ACM-Reference-Format}
\bibliography{
    _ref/ref
}

\clearpage
\appendix
\renewcommand{\thetable}{S\arabic{table}}
\renewcommand{\thefigure}{S\arabic{figure}}

\section*{DIRECT: Supplementary Material}
In this document, we provide additional information including: 
\begin{itemize}[leftmargin=1.5em, topsep=2pt]
\item Details of metrics implementation, human evaluation protocol and dataset specification in Sec.~\ref{supp:details}.
\item Additional ablation study and efficiency analysis in Sec.~\ref{supp:experiments}.
\item Additional qualitative results in Sec.~\ref{supp:example}.
\item Heuristic templates and agent prompts in Sec.~\ref{supp:templates},~\ref{supp:prompts}.
\item Failure cases and limitations discussion in Sec.~\ref{supp:limitations}.

\end{itemize}

\section{Additional Details}
\label{supp:details}

\subsection{Implementation of Explicit Metrics}

\vspace{2pt} 
\noindent \textbf{Semantic - Prompt Relevance ($m_1$).}
This metric assesses the semantic relevance between a selected shot $v_i$ and the user prompt $\mathcal{I}$. We utilize the pre-trained CLIP model (ViT-B/32)~\cite{radford2021clip} as the core representation model. To bridge the domain gap between raw instructional prompts and the descriptive nature of vision-language models, we first employ an LLM to parse the raw prompt $\mathcal{I}$ into a CLIP-optimized text description $\mathcal{I}_{\text{desc}}$, which explicitly encapsulates the desired visual style and thematic content. 
For the visual representation, we sample frames from the shot $v_i$ with a fixed temporal stride of $4$ frames. The unified shot embedding $E(v_i)$ is then derived by calculating the mean of all embeddings across the sampled frames. The final relevance score is defined as the cosine similarity between the semantic embeddings of the LLM-parsed prompt and the aggregated shot: 
$$m_1(v_i, \mathcal{I}) = \cos(E(v_i), E(\mathcal{I}_{\text{desc}}))$$

\vspace{2pt} 
\noindent \textbf{Semantic - Segment Consistency ($m_2$).}
Beyond the prompt relevance of individual shots, maintaining segment-level semantic continuity is essential for a fluid and immersive viewing experience. Frequent or erratic shifts in visual semantics often lead to narrative fragmentation and audience confusion. To quantify this, we evaluate the semantic consistency between consecutive shots $v_{i-1}$ and $v_i$ within the generated sequence. Consistent with the feature extraction protocol defined in $m_1$, we represent each shot using its averaged CLIP embeddings, sampled at a temporal stride of 4 frames. The segment consistency score between two shots is calculated as the cosine similarity of their representations: $$m_2(v_{i-1},v_i) =\cos(E(v_{i-1}), E(v_i))$$

\vspace{2pt} 
\noindent \textbf{Visual - Motion Continuity ($m_3$).}
Professional editing achieves fluid visual transitions between consecutive shots by matching the visual motion direction and velocity. To quantitatively evaluate this, we measure the transition dynamics between the boundary frames of adjacent shots. Using RAFT-Large~\cite{teed2020raft}, we compute the optical flow field $f_1$ between the last two sampled frames (4 frames stride) of the outgoing shot $v_{i-1}$, and $f_2$ between the first two sampled frames of the incoming shot $v_i$. To capture macro-level motion trends, the flow fields are spatially compressed via average pooling.

We decompose the flow similarity into magnitude and direction components. Let $\rho_1$ and $\rho_2$ denote the average vector magnitudes (i.e., motion speed) of $f_1$ and $f_2$, respectively. The magnitude similarity is defined as $s_{mag} = \tau / (|\rho_1 - \rho_2| + \tau)$, and the direction similarity is computed as $s_{dir} = (\cos(f_1, f_2) + 1) / 2$.
Crucially, when the flow magnitude is minimal (e.g., static scenes), the direction similarity becomes heavily influenced by noise and loses reference value. To address this, we introduce a magnitude-aware interpolation strategy. We define a direction confidence weight $w_{dir}$ that penalizes the reliance on $s_{dir}$ when magnitudes are small:
$$w_{dir} = \left( \frac{\rho_1}{\rho_1 + \gamma} \right) \left( \frac{\rho_2}{\rho_2 + \gamma} \right)$$
where $\tau,\gamma$ above are scaling parameters. The final motion continuity score $m_3$ is dynamically interpolated as follows:
$$m_3(v_{i-1}, v_i) = (1 - w_{dir}) \cdot s_{mag} + w_{dir} \cdot s_{dir}$$
This adaptive formulation ensures robust evaluation across both dynamic shots and static scenes.

\vspace{2pt} 
\noindent \textbf{Visual - Framing Consistency ($m_4$).} 
Another professional technique is the use of \textit{match-cutting}~\cite{chen2023match} to overlap subjects' positions between consecutive shots. This technique anchors the viewer's focal point, preventing erratic visual jumps and enabling seamless transitions. To quantify the framing consistency, we evaluate the spatial shift of the visual subject during a transition. 

Specifically, we employ the pre-trained U2-Net~\cite{qin2020u2} to extract the foreground saliency map (a grayscale mask delineating the salient object) from the last sampled frame of the outgoing shot $v_{i-1}$ and the first sampled frame of the incoming shot $v_i$. Rather than relying on naive pixel-wise overlap (e.g, IoU), we normalize these grayscale maps and treat them as 2D spatial probability distributions. This formulation allows us to measure the spatial shift using the 2D Wasserstein Distance, denoted as $\text{dist}(S_{end}(v_{i-1}), S_{start}(v_i))$. Normalizing the distance to $[0, 1]$, the final framing consistency score is computed as:
$$m_4(v_{i-1}, v_i) = 1 - \text{dist}(S_{end}(v_{i-1}), S_{start}(v_i))$$
A higher $m_4$ score indicates a seamless spatial transition where the viewer's attention remains comfortably focused.

\begin{figure}[H]
\centering
  \includegraphics[width=\linewidth]{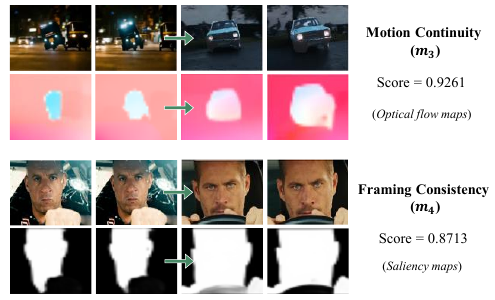}
  \caption{Illustration of Visual Continuity Metrics ($m_3,m_4$).}
  \label{supp:fig:metrics}
\end{figure}

\vspace{2pt} 
\noindent \textbf{Auditory - Beat-Cut Synchronization ($m_5$).} 
A fundamental characteristic of professional video mashups is the precise alignment of visual cuts (shot boundaries) with musical beats, which brings the audience an immersive rhythmic experience.
To evaluate this metric, we first collect the timestamps of all visual cut points $\{t_i\}$ in the video. Concurrently, we employ All-In-One~\cite{kim2023all}, a unified music analyze model to extract the exact timestamps of all musical beats from the background music track $\mathcal{M}$, denoted as $\{b_j\}$.

For each visual cut point $t_i$ (boundary of shot $v_i$), we calculate its temporal distance to the nearest musical beat as $\Delta t_i = \min_j |t_i - b_j|$. The sync score for this cut is then computed as:
$$m_5(v_i, \mathcal{M}) = \exp\left(-\frac{\Delta t_i}{2\sigma^2}\right)$$
where $\sigma$ is a tolerance parameter controlling the decay rate.
Furthermore, to account for potential global audio-visual desynchronization, we enumerate through a temporal offset $\tau \in [-0.5\text{s}, 0.5\text{s}]$ and apply it to all beat timestamps $\{b_j\}$. We calculate the mean sync score across the video for each $\tau$ and report the maximum value as the final score.

\vspace{2pt} 
\noindent \textbf{Auditory - Energy Correspondence ($m_6$).} 
A compelling video mashup synchronizes its visual dynamics with the underlying musical intensity. For instance, high-energy music segments naturally pair with dynamic visual motion, whereas quieter segments suit static or slow-moving shots. To quantify this cross-modal alignment, we evaluate the correlation between visual motion magnitude and auditory energy across the entire video sequence $\mathcal{V}$ consisting of $m$ shots. For each shot $v_i$, we extract its visual motion intensity, $x_i$, defined as the average optical flow magnitude across all sampled frames within the shot. Concurrently, we compute the Root Mean Square (RMS) energy $y_i$ of the music track $\mathcal{M}$ over the exact temporal duration corresponding to $v_i$. This yields two parallel temporal sequences of length $m$: the visual motion sequence $X$ and the auditory energy sequence $Y$. We then compute the Spearman's rank correlation coefficient~\cite{spearman} $\rho(X, Y)$ to measure the monotonic relationship between these two modalities. Finally, to bound the metric within a standard evaluation range, we linearly normalize the correlation coefficient to $[0, 1]$:
$$m_6(\mathcal{V}, \mathcal{M}) = \frac{\rho(X, Y) + 1}{2}$$

\subsection{Human Evaluation Protocol}
We randomly sampled a subset of 10 diverse test cases from \textit{Mashup-Bench}, covering various themes (e.g., action, sci-fi, animation) and musical styles. We invited 60 credible participants, primarily university students from diverse academic backgrounds, to ensure a representative audience. To mitigate cognitive fatigue of long-form video content while maintaining rigorous comparison, each participant was uniformly randomly assigned to two test cases and required to evaluate a triplet of videos for each case (each video approximately 2 minutes in length): (i) \textbf{DIRECT} (Full), (ii) one uniformly random baseline, and (iii) one uniformly random ablation method. For each test case, the videos were anonymized and presented in a randomized order.

Participants were asked to rate each video on a 1-7 Likert scale (1: Very Poor, 7: Excellent) based on three dimensions:

\begin{itemize}[leftmargin=1.5em]
    \item \textbf{Global Structural Alignment}: The video exhibits a cohesive global structure that aligns the narrative arc and visual elements with the overall musical intensity and sections.
    \item \textbf{Local Segment Cohesion}: The visual transitions are fluid with precise beat synchronization, and the synergy between shot semantics and editing styles (e.g., rapid, fluid cuts for action vs. stable long shot for emotional scene) reflects professional editing aesthetics.
    \item \textbf{Overall Quality:} Taking the above factors into account, the mashup video provides an overall compelling, immersive, and professional viewing experience.
\end{itemize}

Before the evaluation, participants were briefed on the high-level objectives and the scoring criteria of video mashups. Subsequently, they viewed each anonymized video in a randomized order, with instructions to focus on the criteria during playback. To ensure the immediacy of feedback, after watching each video, participants provide immediate scores before proceeding to the next one.

\subsection{Dataset and Test Cases Specification}
\begin{figure}[H]
\centering
  \includegraphics[width=\linewidth]{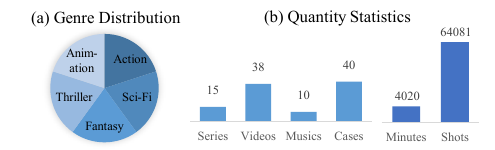}
  \caption{Visualization of Dataset Statistics.}
  \label{supp:fig:dataset}
\end{figure}

To evaluate the robustness of \textbf{DIRECT} across varying cinematic styles, we curated \textit{Mashup-Bench}, a diverse evaluation dataset comprising five distinct genres. The statistics of the dataset is shown in Fig.~\ref{supp:fig:dataset}. Table~\ref{supp:tab:dataset} details the taxonomy of our dataset and provides representative movie examples for each genre.

\begin{table}[ht]
\centering
\caption{Movie Genres of Mashup-Bench.}
\label{supp:tab:dataset}
\small
\begin{tabular}{@{}lll@{}}
\toprule
\textbf{Genre} & \textbf{Movie Example} & \textbf{Key Characteristics} \\ \midrule
Action    & \textit{Mission Impossible}    & Dynamic motion, fast-paced \\
Sci-Fi    & \textit{Interstellar}          & Grand scale, futuristic \\
Fantasy   & \textit{The Lord of the Rings} & Epic narrative, sweeping \\
Thriller  & \textit{Joker}                 & Psychological tension, moody \\
Animation & \textit{Zootopia}              & Vibrant visuals, expressive \\ \bottomrule
\end{tabular}
\end{table}

For each movie series in \textit{Mashup-Bench}, we curated two intra-series test cases with unique music tracks and user prompts. Additionally, we designed two cross-series cases per genre incorporating source footage from multiple movies in this genre. Below are two representative test cases from \textit{Mashup-Bench}.

\newtcolorbox{testcasebox}[1]{
    colback=white,
    colframe=gray!80!black,
    fonttitle=\bfseries,
    title=Input Configuration: #1,
    arc=0mm,
    outer arc=0mm,
    left=5pt,
    right=5pt,
    top=5pt,
    bottom=5pt,
    boxrule=0.8pt
}

\begin{testcasebox}{Cross-Series (Action)}
    \small
    \textbf{Source Footage:} Multiple action movies (Library size: $\sim$10,000 shots) \\
    \textbf{Music Track:} \textit{The Best of Me} (High-tempo, Electronic) \\
    \textbf{User Prompt:} ``Construct a high-intensity adrenaline-pumping video mashup from multiple movie sources. Start with high tension and silence, building up to a chaotic climax with action and chasing. ''
\end{testcasebox}

\vspace{0.5em} 

\begin{testcasebox}{Intra-Series (Thriller)}
    \small
    \textbf{Source Footage:} \textit{Joker} (2019) (Library size: $\sim$2,000 shots) \\
    \textbf{Music Track:} \textit{Play With Fire} (Pop-Rock, Thrilling) \\
    \textbf{User Prompt:} ``Create a stylized montage of \textit{Joker}, focusing on the protagonist's transformation into a chaotic figure. Match the rhythmic, confident beats of the music with the character's erratic movements and scenes of city-wide riots and disorder.''
\end{testcasebox}

\section{Additional Experiments}
\label{supp:experiments}

\subsection{Candidate Shot Sequence Count}
We conduct experiments with varied candidate shot sequence count ($B$) in the Editor module. The quantitative metrics and end-to-end latency of each setting are presented in Tab.~\ref{supp:tab:ablation}. 

\begin{table}[h]
\caption{Ablation on Candidate Shot Sequence Count ($B$).}
\label{supp:tab:ablation}
\centering
\resizebox{\linewidth}{!}{
\begin{tabular}{@{}lccccccc@{}}
\toprule
\textbf{Count} ($B$) & \textbf{$m_1 \uparrow$} & \textbf{$m_2 \uparrow$} & \textbf{$m_3 \uparrow$} & \textbf{$m_4 \uparrow$} & \textbf{$m_5 \uparrow$} & \textbf{$m_6 \uparrow$} & \textbf{Latency} (s) $\downarrow$ \\ \midrule
$B=1$ (Greedy) & 25.22 & 84.78 & 75.10 & 82.36 & \textbf{98.73} & 80.24 & \textbf{380} \\
$B=3$ (Ours)   & 25.49 & 84.97 & 77.26 & 83.37 & 98.69 & 82.40 & 646 \\
$B=5$          & \textbf{25.54} & \textbf{85.36} & \textbf{77.95} & \textbf{83.61} & 98.70 & \textbf{83.15} & 891 \\ \bottomrule
\end{tabular}
}
\end{table}

As shown, a greedy approach ($B=1$) minimizes latency but yields suboptimal performance due to its limited search range. Conversely, increasing the candidate count to $B=5$ yields minor improvements in explicit metrics and greater flexibility for validation choices, but incurs a substantial increase in both search and validation time. Therefore, we empirically set the candidate shot sequence count to $B=3$ to strike an optimal balance between generation quality and computational efficiency.

\subsection{Efficiency Analysis}
We analyze the latency breakdown of the DIRECT framework to evaluate its overall computational efficiency. 

\begin{figure}[H]
\centering
  \includegraphics[width=\linewidth]{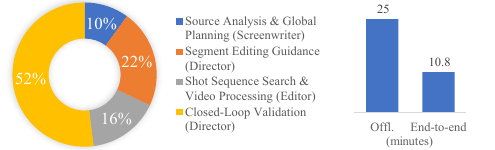}
  \caption{Latency Breakdown and Average Time Cost.}
  \label{supp:fig:efficiency}
\end{figure}

As shown in Fig.~\ref{supp:fig:efficiency}, the primary computational bottleneck resides in the MLLM reasoning processes, particularly during the closed-loop validation phase which requires analyzing multiple candidate shot sequences. In contrast, the Editor's sequence search is efficient and accounts for a small fraction of the total runtime, as it directly leverages offline pre-computed features rather than relying on real-time neural network inference. This validates our design choice of offline feature extraction, successfully decoupling heavy visual perception from the online iterative editing loop.

\subsection{MLLM Backbones}
To evaluate the capabilities of different MLLM backbones, we conduct a qualitative case study comparing the intermediate agent outputs of Qwen3-VL-8B-Instruct and a larger commercial model, GPT-4o. As the case study below reveals, while GPT-4o occasionally utilizes a more expressive vocabulary in its reasoning rationales, both models produce highly effective outputs that successfully fulfill the core objectives of global structural alignment and local segment cohesion. This highlights the robustness of our hierarchical task decomposition, proving Qwen3-VL-8B-Instruct to be a highly efficient, fully reproducible, and functionally capable backbone.

\newtcblisting{agentbox}[1]{
    enhanced,
    breakable,
    colback=white,
    colframe=gray!80!black,
    title=\textbf{#1},
    boxrule=0.5pt,
    arc=0mm,
    outer arc=0mm,
    fonttitle=\small\sffamily,
    left=4pt, right=4pt, top=4pt, bottom=4pt,
    listing only, 
    listing options={
        style=tcblatex,
        basicstyle=\footnotesize\ttfamily,
        breaklines=true,
        breakautoindent=false,
    }
}

\begin{agentbox}{Case Study: Qwen3-VL-8B-Instruct Intermediate Output}
[Screenwriter Section Plan]
{
  "section_name": "chorus",
  "energy_level": "Extreme",
  "visual_tags": ["explosion", "wide chaos", "vehicular mayhem", "dynamic", "fast-paced"],
  "rationale": "Climactic section with maximum visual intensity, synchronized with music’s heaviest beat count and highest dB level."
}

[Director Segment Guidance]
{
  "thought_process": "1. Analyze Vibe: Chorus with extreme energy... 3. Simplify query: 'Vehicle crashing violently' captures the destructive energy without repetition.",
  "retrieval_query": "Vehicle crashing violently"
}
{
  "thought_process": "The query implies high-energy, chaotic motion... making Motion_Continuity_Priority ideal for preserving momentum and impact.",
  "weight_profile": "Motion_Continuity_Priority"
}
{
  "thought_process": "1. Analyze Content: 'Vehicle crashing violently' is high-energy... 3. Length check: 12 beats fits within 24 beats remaining.",
  "pacing_control": [2, 2, 2, 2, 2, 2]
}


\end{agentbox}

\begin{agentbox}{Case Study: GPT-4o Intermediate Output}
[Screenwriter Section Plan]
{
  "section_name": "chorus",
  "energy_level": "Maximum",
  "visual_tags": ["extreme speed", "rapid intercut", "engine sync", "collision chaos", "intense pursuit"],
  "rationale": "Climactic sequence with relentless pacing, syncing engine actions and cuts tightly to percussion."
}

[Director Segment Guidance]
{
  "thought_process": "1. Vibe is maximum energy... focus on intense vehicular action. 2. Depart from previous queries, choose urban chase cluster. 3. Focus: Speed and danger.",
  "retrieval_query": "car crashing at high speed"
}
{
  "thought_process": "Query focuses on a high-speed car crash... For maximum energy and rapid intercut rhythm, Motion_Continuity_Priority is key to preserving kinetic flow.",
  "weight_profile": "Motion_Continuity_Priority"
}
{
  "thought_process": "1. Analyze Content: High-intensity action... 2. Strategy: Use very short cuts (1-2 beats) to match rapid intercut. 3. Length check: 10 beats fits within 24 remaining.",
  "pacing_control": [1, 1, 1, 1, 2, 1, 1, 2]
}
\end{agentbox}

\section{Additional Qualitative Results}
\label{supp:example}

\begin{figure*}[ht] 
  \centering
  \includegraphics[width=\textwidth]{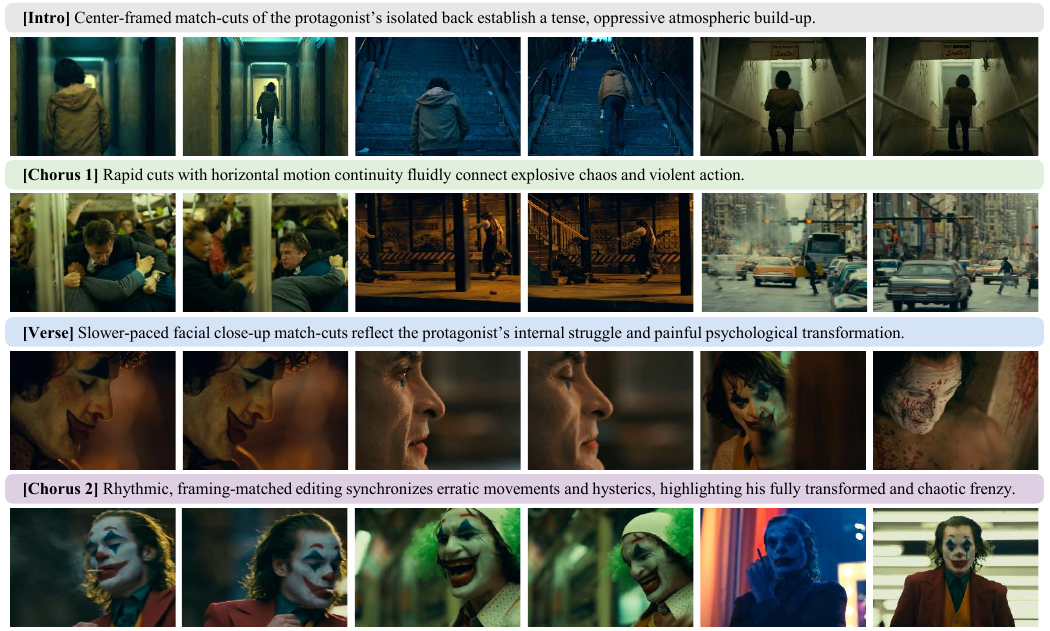}
  \caption{Representative shot sequences for each section of a complete video result. The sequences demonstrate the global structural alignment and local segment cohesion achieved by our framework.}
  \label{supp:fig:example}
\end{figure*}

We provide additional qualitative results featuring representative shot sequences for each section of a complete video result, as shown in Fig.~\ref{supp:fig:example}. These sequences further demonstrate the effectiveness of our framework in achieving global structural alignment and local segment cohesion.

\section{Predefined Heuristic Templates}
\label{supp:templates}

As shown in Tab.~\ref{supp:tab:templates}, we define five heuristic templates to reflect varying editing priorities under different scenarios. These templates determine the balancing weights ($w_{heu}$) for the low-level metrics during the shot sequence search. The weights are empirically tuned to ensure the generated segments optimally align with the intended visual and rhythmic styles.

\begin{table*}[ht]
    \centering
    \caption{Predefined heuristic templates ($\mathbf{w}_{heu}$).}
    \label{supp:tab:templates}
    \begin{tabular}{@{}llc@{}}
    \toprule
    \textbf{Template Name} & \textbf{Suitable Scenarios} & \textbf{Prioritization} \\ \midrule
    \texttt{<Semantic\_Priority>} & Narrative focus or specific subjects & stricter shot pool, $m_2$ \\
    \texttt{<Motion\_Continuity\_Priority>} & Fluid action sequences (e.g., chases, racing) & $m_3,m_6$ \\
    \texttt{<Composition\_Similarity\_Priority>} & Static framing or match-cutting (e.g., close-ups) & $m_4$ \\
    \texttt{<Hybrid\_Visual\_Coherent>} & Intricate action balancing motion and framing & $m_3, m_4$ \\
    \texttt{<Default\_Priority>} & Balanced, general-purpose retrieval & - \\ \bottomrule
    \end{tabular}
\end{table*}

\section{MLLM Agents and System Prompts}
\label{supp:prompts}

This section provides the system prompts for the agentic modules in the \textbf{DIRECT} framework. To ensure conciseness, we present the \textit{abbreviated} prompts (omitting input templates and in-context learning examples) that define the operational logic, role-specific guidelines, and the interaction protocol for the MLLM agents.

\lstnewenvironment{promptbox}{
    \lstset{
        basicstyle=\footnotesize\ttfamily,
        columns=flexible,
        breaklines=true,      
        breakindent=0pt,  
        breakautoindent=true,     
        frame=single,           
        framerule=0.8pt,    
        rulecolor=\color{gray!30},   
        framesep=8pt,      
        xleftmargin=2pt,    
        xrightmargin=2pt,    
        keepspaces=true,        
        showstringspaces=false,   
    }
}{}

\subsection{Screenwriter (Cluster Captioning)}

\begin{promptbox}
# Role
Footage Archivist Agent: Specialized in visual abstraction and semantic clustering for cinematic content.

# Task
Analyze a cluster of {n_shots} semantically similar movie shots to identify their "Common Visual Archetype." Create a generic, high-level description that defines the core invariant features of the cluster.

# Operational Guidelines
1. De-Specify: Strictly remove proper nouns, specific actors, brand names, or real-world locations.
2. Generalize: Abstract specific entities into broad categories (e.g., use "vehicle" instead of "sports car").
3. Visual Invariants: Identify the constant elements across Subject, Action, Background, and Shot Type.
4. Vibe Extraction: Distill the environmental aesthetic and atmospheric tone into concise keywords.

# Output Format
- Visual Archetype: [The simplified common archetype description]
- Visual Vibe: [3 keywords]
\end{promptbox}

\subsection{Screenwriter (Summary Synthesis)}

\begin{promptbox}
# Role
Footage Library Analyst: A high-level synthesizer for intelligent video editing systems, specialized in cinematic trend extraction and library-wide metadata analysis.

# Task
Process a corpus of {total_shots} visual archetypes to generate a "Footage Analysis Report." Synthesize discrete cluster data into a unified visual identity for editorial direction.

# Operational Guidelines
1. Aesthetic Deduction: Infer the global visual style (e.g., Gritty/Handheld vs. Static/Composed) from the collective metadata.
2. Keyword Distillation: Extract 10-15 high-impact cinematic keywords defining the library's visual DNA.
3. Thematic Clustering: Categorize raw descriptions into 8-10 prominent "Cinematic Themes" suitable for highlight montages.
4. Descriptive Precision: For each theme, provide a concise title and a two-sentence definition:
   - Sentence 1: General subject and action (de-specified).
   - Sentence 2: Defining visual features (vibe, shot type, environment).

# Target Output Format
Footage Analysis Report: [Title]
- Visual Style & Tone: [Overall Aesthetic Deduction]
- Global Keywords: [10-15 Keywords]
- Visual Clusters for Retrieval: 
  - [Theme Title]: [Two-sentence definition]
  - ...
\end{promptbox}

\subsection{Screenwriter (Structural Plan)}

\begin{promptbox}
# Role
Video Mashup Planner: An expert in cinematic rhythm and narrative synchronization, specialized in mapping visual energy to musical structures.

# Task
Synthesize [User Prompt], [Footage Summary], and [Music Profile] into a dual-layered production plan:
1. Global Narrative Flow: A high-level emotional and story arc.
2. Detailed Section Plan: A synchronized JSON mapping for every music segment.

# Operational Guidelines
1. Energy-Sync: Align visual intensity (Shot Type, Action) with musical energy (e.g., Intro=Low, Chorus=High).
2. Multi-Dimensional Tagging: Each section must include Subject/Action, Atmosphere/Vibe, Shot Type, and Energy Level.
3. Tag Generality: Use generic descriptors (e.g., "intense fighting") instead of specific scenarios to maximize retrieval success.
4. Contrast & Diversity: Ensure visible variance in tags between adjacent sections to reflect musical transitions.
5. Completeness: Generate exactly one JSON object for every music section provided in the input.

# Output Format
## Global Narrative Flow
[Describe the overall story arc, emotional build-up, and climax.]

## Detailed Section Plan (Strictly JSON)
[
  {
    "section_name": "Section Title",
    "energy_level": "Low/Medium/High",
    "visual_tags": ["tag1", "tag2", "tag3", "tag4"],
    "rationale": "Explanation of alignment with music/theme"
  },
  ...
]
\end{promptbox}

\subsection{Director (Semantic Query)}

To tackle the complex task of editing intent instantiation and guidance translation, we take advantage of the Chain-of-Thoughts prompting technique to generate the segment editing guidance step by step.

\begin{promptbox}
# Role
Director Agent (Semantic Query): A retrieval specialist for intelligent video mashup systems, focused on translating section vibe and keywords into CLIP-friendly visual queries.

# Task
Generate a concise "Retrieval Query" (2-7 words) for the current video segment. The query must align with the music section's energy and keywords while ensuring a high probability of successful footage retrieval.

# Operational Guidelines
1. Vibe & Energy Alignment: Match the visual intensity and thematic tone to the current music section's energy level and keywords.
2. CLIP Optimization: Use simple "Subject + Action/Description" structures. Avoid complex adjectives; focus on core, retrievable visual elements to ensure a wide search range.
3. Diversity Protocol: Review the previous 4 segments' queries to ensure visual variety. Avoid repeating the same subjects or actions to maintain montage dynamism.
4. Error Adaptation: If a "Prior Failure" is provided, analyze the feedback and further generalize the description or switch to a different subject within the same vibe to resolve the rejection.

# Output Format (Strictly JSON)
{
  "thought_process": "1. Analyze vibe/energy. 2. Verify history to avoid repetition. 3. Simplify to a 2-7 word retrieval string.",
  "retrieval_query": "Final CLIP-friendly query"
}
\end{promptbox}

\subsection{Director (Editing Heuristics)}

\begin{promptbox}
# Role
Director Agent (Optimization Strategy): A technical coordinator specialized in parameter tuning for visual retrieval and shot-selection optimization.

# Task
Analyze the "Target Visual Query" from the previous phase and select the most effective "Constraint Weight Profile" to ensure high-fidelity video segment retrieval from the footage library.

# Operational Guidelines
1. Kinetic Assessment: Evaluate the query's motion requirements (e.g., high-speed action vs. static portraiture) and cinematic complexity.
2. Profile Mapping: Select exactly one profile from the predefined technical library:
   - Semantic_Priority: For narrative focus or specific subjects.
   - Motion_Continuity_Priority: For fluid action sequences (e.g., chases, racing).
   - Composition_Similarity_Priority: For static framing or match-cutting (e.g., close-ups).
   - Hybrid_Visual_Coherent: For intricate action balancing motion and framing (e.g., combat).
   - Default_Priority: For balanced, general-purpose retrieval.
3. Contextual Integration: Factor in the current music section's energy level to prioritize either temporal flow or visual detail.

# Output Format (Strictly JSON)
{
  "thought_process": "Analysis of query kinetics, musical energy alignment, and profile selection logic.",
  "weight_profile": "Profile_Name"
}
\end{promptbox}

\subsection{Director (Rhythmic Pacing)}

\begin{promptbox}
# Role
Director Agent (Rhythmic Execution): A temporal pacing specialist for intelligent montage systems, responsible for mapping visual cuts to musical meter and energy.

# Task
Calculate the precise "Pacing Control" (a sequence of shot durations in beats) for the current segment. The output must be a list of integers whose sum fits within the available musical section duration.

# Operational Guidelines
1. Temporal Constraints: A reasonable total segment duration is within 4-16 beats and must strictly $\le$ beats_remaining.
2. Musical Alignment:
   - High Energy: Utilize short durations (1-2 beats) for rapid-fire editing.
   - Low Energy: Utilize long durations (4, 8, 16 beats) for sustained shots.
   - Triple Meter (3/4): Prioritize 3 or 6-beat cuts to maintain metric synchronicity.
3. Adaptive Shot Density:
   - High Density (Common/Action): For abundant footage (e.g., running, fighting), use frequent cuts to increase visual dynamism.
   - Low Density (Specific/Emotional): For rare or narrative-heavy footage (e.g., crying, explosions), use fewer, longer shots to preserve detail.
   - Medium Density (Atmospheric): For landscapes or establishing shots, use balanced 4-8 beat durations to allow the visual to breathe.

# Output Format (Strictly JSON)
{
  "thought_process": "1. Analyze content density. 2. Align with energy/meter. 3. Final length check against beats_remaining.",
  "pacing_control": [2, 2, 2, 2]
}
\end{promptbox}

\subsection{Director (Validation)}

\begin{promptbox}
# Role
Quality Assessment Specialist: A senior film editor specialized in evaluating video candidates from beam search results, focusing on narrative alignment, rhythmic pacing, and visual continuity.

# Task
Evaluate a batch of video candidates against the "Retrieval Query" and "Guidance Profile." Select the optimal candidate for the current segment or reject the entire batch if no candidate meets the minimum cinematic standard.

# Operational Guidelines
1. Semantic Alignment: Verify if the candidate matches the core subject and action of the retrieval query (apply reasonable tolerance for non-literal matches).
2. Structural Integrity: Check for visual coherence, avoiding technical glitches, jarring transitions, or artifacts.
3. Selection Pragmatism: Prioritize selection over rejection unless candidates are fundamentally unrelated to the query or aesthetically broken.
4. Error Feedback: If rejecting all candidates, provide specific diagnostic reasons and actionable suggestions for query re-generation or subject switching.

# Output Format (Strict JSON)
{
    "success": boolean,           // False if all candidates fail critical criteria
    "best_candidate": int | null, // 0-based index of the chosen clip
    "verdict": "string",          // Concise professional summary of the decision
    "issues": ["string"],         // Specific flaws identified in candidates
    "suggestions": ["string"]     // Feedback for the Director if success is false
}
\end{promptbox}

\section{Failure Cases and Limitations}
\label{supp:limitations}

While the \textbf{DIRECT} framework demonstrates superior performance in creating professional-grade video mashups, we acknowledge two primary limitations in our current implementation.

\begin{itemize}[leftmargin=1.5em, topsep=2pt]
    \item \textbf{Agent Hallucination and Retrieval Failure:} Despite our closed-loop validation designed to dynamically relax semantic constraints, MLLM agents occasionally generate highly specific queries absent from the available library. Exhausting the retry limit forces the system to fall back on suboptimal sequences. This highlights the need for more robust, grounding-aware footage summarization for large-scale heterogeneous libraries in future works.
    
    \item \textbf{Perception Bottleneck of Feature Extractors:} The Editor heavily relies on pre-trained foundational models for low-level visual features. These models often degrade under extreme visual conditions (e.g., severe motion blur or dim lighting) or exhibit domain gaps on highly stylized footage. Consequently, the Editor may inadvertently optimize for visual artifacts rather than true coherency, revealing a bottleneck in applying general-purpose models to diverse cinematic domains.
\end{itemize}

\end{document}